\documentclass{article}

\usepackage{microtype}
\usepackage{graphicx}
\usepackage{subfigure}
\usepackage{subcaption}
\usepackage{subcaption}
\usepackage{booktabs} 
\usepackage{algorithm}

\usepackage{hyperref}
\usepackage{times}
\usepackage{latexsym}
\usepackage{multirow}
\usepackage{tikz}
\usepackage{pgfplots}
\pgfplotsset{compat=1.18}
\usepackage{subcaption}
\usepackage{amsmath} 
\usepackage{amssymb}
\usepackage{graphicx}
\usepackage{tikz}
\usetikzlibrary{arrows.meta,positioning}
\usetikzlibrary{positioning, calc, arrows.meta, shapes.multipart, fit}
\usetikzlibrary{arrows.meta,positioning}
\usepackage{xcolor}
\usetikzlibrary{positioning}
\usetikzlibrary{calc}
\hypersetup{hidelinks}
\usepackage{enumitem}
\usepackage{booktabs}      
\usepackage{tabularx}      
\usepackage[most]{tcolorbox}
\usepackage{xcolor}        
\tcbuselibrary{skins}      
\usepackage{bm}

\usepackage[T1]{fontenc}

\usepackage[utf8]{inputenc}

\usepackage{microtype}

\usepackage{inconsolata}

\usepackage{graphicx}

\usepackage[accepted]{icml2025}

\usepackage{amsmath}
\usepackage{amssymb}
\usepackage{mathtools}
\usepackage{amsthm}

\usepackage[capitalize,noabbrev]{cleveref}

\theoremstyle{plain}

\theoremstyle{definition}

\theoremstyle{remark}

\usepackage[textsize=tiny]{todonotes}
\icmltitlerunning{}

\icmltitlerunning{Learning to Refine Hidden States for Reliable LLM Reasoning}

\begin{document}

\twocolumn[
\icmltitle{Learning to Refine Hidden States for Reliable LLM Reasoning}

\begin{icmlauthorlist}
\icmlauthor{Chia-Hsuan Hsu}{}
\icmlauthor{Jui-Ming Yao}{}
\end{icmlauthorlist}

\vskip 0.3in
]




\begin{abstract}
Large language models have achieved strong performance on reasoning and generation tasks, 
but reliable reasoning remains challenging in complex multi-step settings. In such settings, 
models must integrate multiple pieces of evidence, track intermediate conclusions, and 
maintain consistency across successive reasoning steps. Existing explicit reasoning methods, 
such as chain-of-thought prompting, improve performance by generating intermediate 
rationales, but they do not directly regulate the model's internal reasoning process and 
often introduce substantial inference overhead. We propose \textbf{ReLAR}, a 
reinforcement-guided latent refinement framework that iteratively updates hidden 
representations before decoding. ReLAR maintains a compact latent reasoning state and 
uses learned depth and action controllers to adaptively determine both the number and 
direction of refinement steps. The controllers are trained with a policy-gradient objective 
based on step-wise likelihood improvement, enabling input-dependent reasoning control 
without relying on explicit reasoning. Experiments on medical, mathematical, multi-hop 
reasoning, and open-ended generation benchmarks show that ReLAR improves task 
performance, generation quality, and reasoning stability while requiring substantially lower 
inference overhead than explicit reasoning baselines. Code is available at 
\href{https://github.com/tongyu0924/Learning-to-Refine-Hidden-States-for-Reliable-LLM-Reasoning}
{\texttt{tongyu0924/\allowbreak Learning-to-Refine-\allowbreak Hidden-States}}.
\end{abstract}

\section{Introduction}
Large language models (LLMs) have demonstrated strong capabilities across a
wide range of reasoning and generation tasks, including question answering,
mathematical problem solving, multi-hop reasoning, clinical summarization, and
open-ended text generation~\cite{singhal2023large,thirunavukarasu2023large,lucas2024ensemble}.
Despite these advances, reliable reasoning remains difficult in complex
multi-step settings, where models must integrate multiple pieces of evidence,
maintain intermediate conclusions, and preserve consistency across successive
reasoning steps. Even small errors can propagate through later computations and
lead to incorrect conclusions~\cite{chen2025evalhealthcare,he2025survey}.

A key challenge is that such failures are often not caused solely by missing
knowledge, but by instability in the model's internal reasoning state. During
multi-step reasoning, a model may overemphasize a salient but misleading cue,
underweight relevant evidence, or gradually drift toward an inconsistent
intermediate conclusion. Improving reasoning reliability therefore requires
mechanisms that can stabilize and control the internal representations from
which answers are generated.

The dominant approach for eliciting reasoning is explicit reasoning, such as
chain-of-thought (CoT) prompting, which asks the model to generate intermediate
reasoning steps in natural language
~\cite{wei2022chain,wang2022selfconsistency,yao2023tree,shinn2023reflexion}.
Although effective, these methods operate primarily at the level of generated
text and do not directly regulate the hidden representations that support the
model's internal reasoning process. Moreover, textual reasoning traces may
contain logical gaps or hallucinated content even when final answers appear
fluent or correct~\cite{lyu2023faithfulcot,lanham2023faithfulness}, while long
reasoning traces increase inference latency and computational cost.

\begin{figure*}[t]
  \centering
  \includegraphics[width=0.8\textwidth]{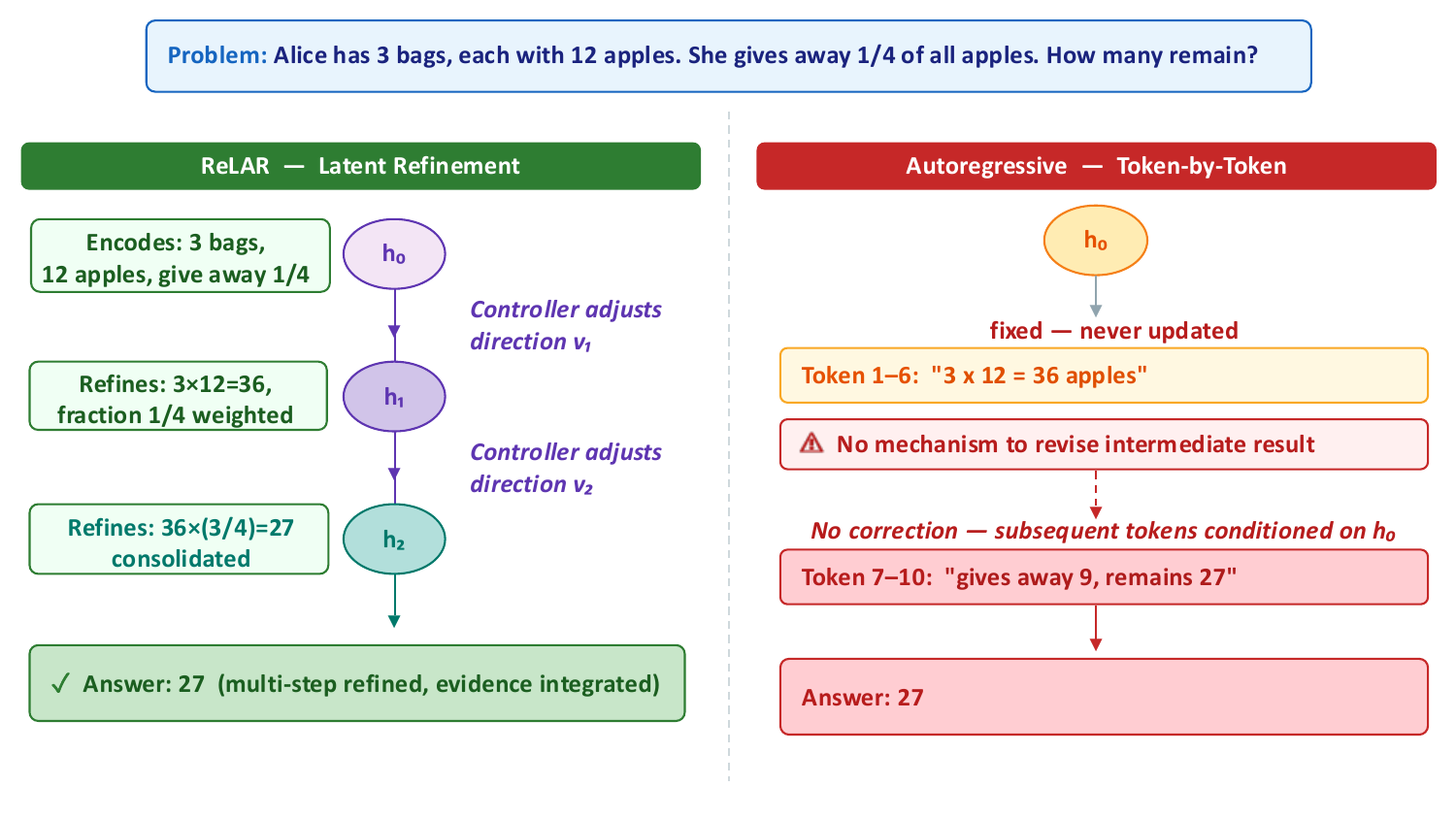}
  \caption{Comparison of ReLAR and conventional autoregressive reasoning.
ReLAR iteratively refines the hidden representation before decoding, whereas conventional autoregressive decoding proceeds through token-level generation without an explicit pre-decoding latent refinement loop. The example illustrates the difference in computation structure rather than a failure case of autoregressive decoding.}
  \label{fig:math_comparison}
\end{figure*}

These limitations motivate an alternative question: instead of externalizing
reasoning as a sequence of text tokens, can we improve reasoning reliability by
refining the model's internal hidden states before decoding? As illustrated in
Figure~\ref{fig:math_comparison}, conventional autoregressive decoding proceeds
through token-level generation without an explicit pre-decoding latent
refinement loop. In contrast, latent refinement performs intermediate
computation directly in hidden-state space before output tokens are produced.

Recent studies on latent representation editing and activation intervention
show that hidden states encode semantically meaningful information and that
intervening on internal activations can influence model behavior more directly
than output-level supervision
~\cite{wang2025semantics,stolfo2025improving,helff2026activationreasoning,meng2022rome}.
However, most existing latent intervention methods rely on static, single-step,
or task-specific mechanisms, and therefore do not support adaptive multi-step
refinement of internal reasoning states.

To address this gap, we propose ReLAR, a reinforcement-guided latent refinement
framework that performs multi-step, learnable refinement directly in hidden-state
space before decoding. Rather than generating explicit reasoning traces, ReLAR
iteratively updates the model's internal representation, allowing the reasoning
state to be progressively adjusted and stabilized prior to answer generation.
Two learned controllers dynamically determine the refinement direction and the
number of refinement steps, enabling input-dependent allocation of latent
reasoning computation according to task difficulty.

Our contributions can be summarized as follows:
\begin{enumerate}[leftmargin=*]
\item We propose ReLAR, an alternative to explicit chain-of-thought reasoning
that performs multi-step, learnable refinement directly in hidden-state
space before decoding. This allows the model to adjust and stabilize its
internal reasoning process without incurring the cost of generating
intermediate textual rationales.

\item We introduce reinforcement-learning-based depth and action controllers
that dynamically determine how many refinement steps to perform and in
which direction each hidden-state update should move. This enables
input-dependent, learnable, and adaptive allocation of internal reasoning
computation.

\item Experiments across medical, mathematical, multi-hop reasoning, and
open-ended generation benchmarks demonstrate that ReLAR improves
accuracy, generation quality, and reasoning stability while requiring
substantially lower inference-time overhead than explicit reasoning-based
baselines such as chain-of-thought prompting.
\end{enumerate}

\section{Related Work}

\subsection{Implicit Reasoning in Large Language Models}

Large language models (LLMs) can perform complex reasoning not only through
explicit natural-language rationales, but also through implicit computation
within their internal representations. While chain-of-thought prompting
elicits intermediate reasoning steps in
text~\citep{wei2022chain,kojima2022large}, recent studies suggest that models
may encode task-relevant reasoning information in hidden states even when such
reasoning is not explicitly
verbalized~\citep{schlag2021linear,geva2021transformer,hao2024training,saunshi2025reasoning}.

Implicit reasoning is attractive because it avoids the computational cost and
potential unfaithfulness of long textual rationales, while still allowing the
model to integrate evidence before producing an answer. Recent latent-reasoning
approaches further explore this direction by performing intermediate
computation in continuous hidden-state space rather than through generated
tokens~\citep{hao2024training,saunshi2025reasoning,geiping2026scaling}.
However, standard LLM inference usually relies on a single forward pass,
leaving the implicit reasoning process largely uncontrolled. Our work builds
on this view by treating reasoning as an internal latent process that can be
iteratively refined before generation.

\subsection{Latent Reasoning and Representation-Level Refinement}

Building on latent-reasoning studies, recent work investigates whether model
behavior can be controlled by directly manipulating hidden representations.
Such approaches include hidden-state editing, activation steering, and
activation-space reasoning, which intervene on internal activations rather than
on generated rationales~\citep{elazar2021amnesic,geva2021transformer,
meng2022rome,wang2025semantics,stolfo2025improving,
helff2026activationreasoning}. These studies show that hidden representations
contain semantically meaningful information and can be used to influence model
outputs before or during decoding.

However, existing representation-level methods often rely on fixed,
single-step, or task-specific intervention mechanisms. Although they provide
useful control over model behavior, they offer limited support for adaptive
multi-step refinement of internal reasoning states. In contrast, ReLAR treats
hidden-state refinement as an iterative control process, learning to
adaptively select both the refinement direction and depth before decoding.

\subsection{Reinforcement Learning for Adaptive Reasoning Control}

Reinforcement learning (RL) has been widely used in large language systems for
policy optimization, preference alignment, reward shaping, and reward-guided
decision making~\citep{ouyang2022training,bai2022training,rafailov2023direct}.
These methods demonstrate that learned policies can optimize model behavior
under feedback signals rather than relying solely on supervised next-token
prediction.

For reasoning tasks, RL provides a natural framework for adaptive control,
where a model can learn to select actions based on the current reasoning state
and expected future improvement. Recent work has explored RL for improving
reasoning through verifiable rewards, self-correction, and tool-using
agents~\citep{kumar2024training,xie2025logic,feng2025retool,
yuan2026verifiable}. However, these methods typically operate through
generated text, external tool actions, or output- and process-level feedback,
rather than directly controlling the model's latent reasoning process.

Our work applies RL to hidden-state refinement itself. ReLAR trains dedicated
controllers to decide both how many refinement steps to perform and how each
latent update should be directed. This allows adaptive reasoning control to
occur before decoding, directly within the model's internal representation
space.

\begin{figure*}[t]
    \centering
    \includegraphics[width=\textwidth]{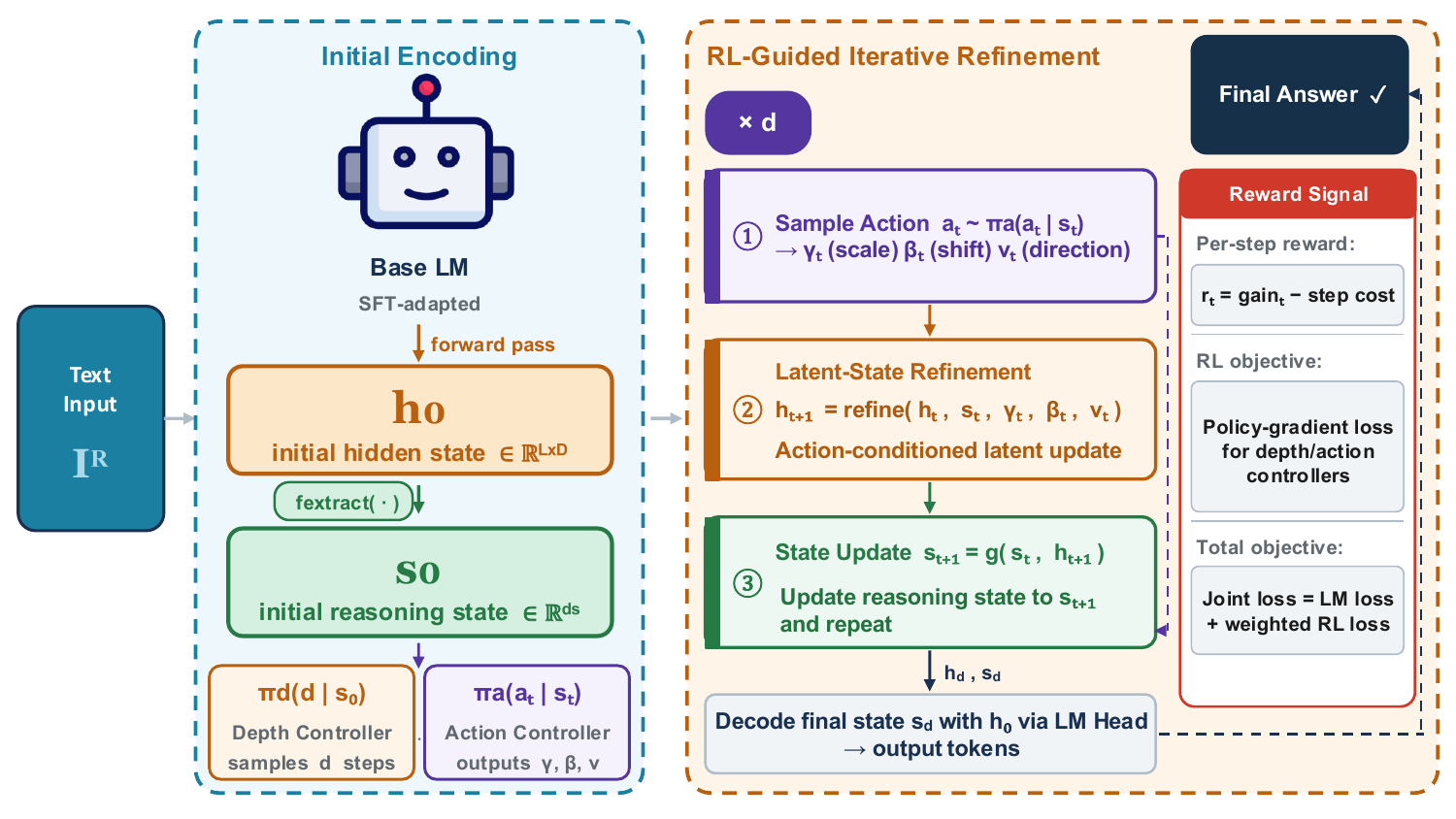}
    \caption{Overview of the model pipeline.}
    \label{fig:pipeline}
\end{figure*}

\begin{figure}[!htbp]
    \centering
    \includegraphics[width=\columnwidth]{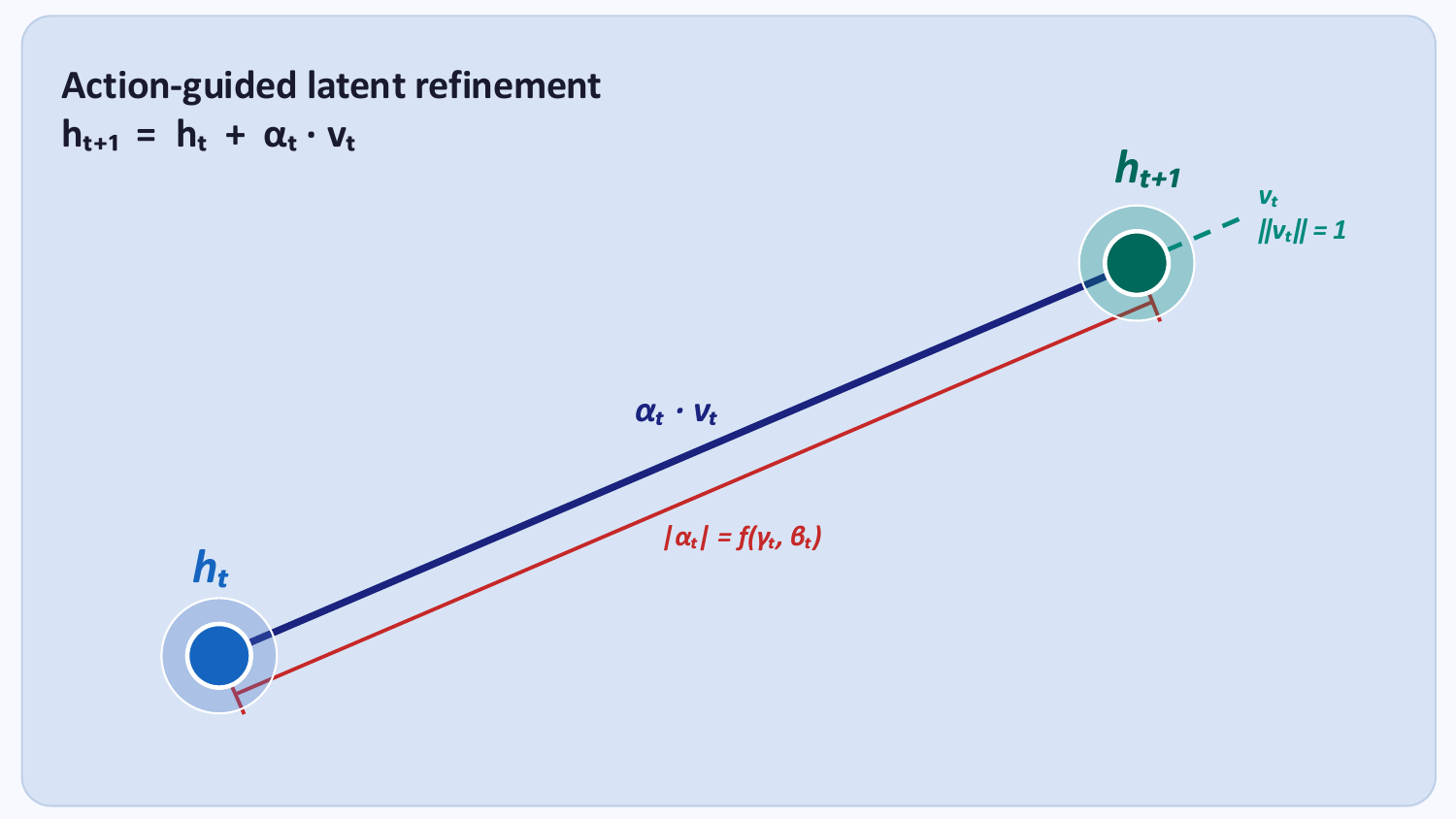}
    \caption{Iterative latent-state refinement}
    \label{fig:latent_refinement}
\end{figure}

\section{Methodology}
\label{sec:method}

We introduce \textbf{ReLAR (Reinforcement-Guided Latent Refinement)}, an
iterative hidden-state refinement framework that enables controllable,
multi-step reasoning entirely within the latent space of a pretrained language
model. Rather than producing an answer from a single forward pass, ReLAR
executes a sequence of representation-refinement steps before decoding, guided
by two learned controllers that adaptively determine \emph{how deeply} and
\emph{in which direction} the hidden state should be revised. This design
allows the model to perform input-dependent latent computation before
generation, so that internal evidence can be progressively adjusted and
consolidated without producing explicit chain-of-thought tokens.
Figure~\ref{fig:pipeline} gives an overview of the full pipeline.

We provide additional methodological and reproducibility details in the
supplementary material, including the complete training procedure in
Algorithm~\ref{alg:relar_training} in Appendix~\ref{apd:algorithm},
the prompt templates in Appendix~\ref{apd:prompt_templates},
the training configuration, value critic details, refinement direction
projection, answer extraction, dataset processing, baseline evaluation,
random seeds and released artifacts, and the notation summary in
Section~\ref{sec:notation}.

\subsection{Preliminaries}
\label{sec:prelim}

Let $x = (x_1, \dots, x_n)$ be an input token sequence and let $p_{\theta}$
denote a transformer-based language model. A single forward pass yields a
final-layer hidden state $h_0 \in \mathbb{R}^{L \times D}$, where $L$ is the
sequence length and $D$ is the hidden dimension.

Our framework augments $p_{\theta}$ with a low-dimensional \textit{reasoning
state} $s_t \in \mathbb{R}^{d_s}$ that tracks the model's evolving internal
latent representation. Unlike chain-of-thought rationales, $s_t$ is never
decoded into text. Starting from $(h_0, s_0)$, the framework produces coupled
refinement trajectories
\[
  s_0 \to s_1 \to \cdots \to s_T,
  \qquad
  h_0 \to h_1 \to \cdots \to h_T,
\]
where $T$ is selected adaptively by a learned depth controller.

\subsection{Initial Reasoning State}
\label{sec:init}

Given the base hidden representation $h_0$, we construct the initial reasoning
state via a lightweight projection network $f_{\mathrm{extract}}$:
\[
  s_0 = f_{\mathrm{extract}}(h_0).
\]
The vector $s_0 \in \mathbb{R}^{d_s}$ compresses task-relevant information
from the final transformer layer into a compact representation. This compact
state provides an efficient interface for refinement control, avoiding the need
for the controllers to operate directly on the full token-level hidden state.
It acts as a \emph{shared bottleneck}: both the depth controller $\pi_d$ and
the action controller $\pi_a$ condition on this state when determining the
refinement depth and update direction.

\subsection{Action-Guided Representation Refinement}
\label{sec:refine}
At each refinement step $t \in \{0,\ldots,T-1\}$, the action controller
predicts a structured latent update rather than directly replacing the hidden
state. Specifically, it outputs a refinement direction and two modulation
parameters:
\[
a_t = (\gamma_t,\beta_t,\bm{v}_t) \sim \pi_a(a_t \mid s_t),
\]
where $\bm{v}_t \in \mathbb{R}^{d_v}$ is normalized to satisfy
$\|\bm{v}_t\|=1$. Concretely, $\pi_a$ parameterizes $\gamma_t,\beta_t,\bm{v}_t$
as independent Gaussians conditioned on $s_t$; $\bm{v}_t$ is subsequently
projected onto the unit sphere, and $\log \pi_a(a_t\mid s_t)$ is computed from
the pre-projection Gaussian densities. The direction vector $\bm{v}_t$
determines where the hidden representation should move in latent space, while
the scalar parameters $\gamma_t$ and $\beta_t$ control the magnitude and sign
of the update through an effective signed step size
$\alpha_t = f_{\alpha}(\gamma_t,\beta_t)$.
The hidden representation is then refined by an additive update:
\[
h_{t+1} = h_t + \alpha_t \bm{v}_t.
\]
As illustrated in Figure~\ref{fig:latent_refinement}, each refinement step
moves the hidden representation along the normalized direction $\bm{v}_t$ with
the signed step size $\alpha_t$. This local update keeps the refined
representation anchored to the previous state, which avoids generating an
unconstrained hidden-state replacement at each step. After applying the update,
the reasoning state is refreshed:
\[
s_{t+1}=g(s_t,h_{t+1}).
\]
This allows the controller to condition the next action on the effect of the
previous refinement step.
Rather than decoding from the accumulated trajectory $h_1,\ldots,h_T$
directly, we anchor the final representation to the original $h_0$ to
preserve the full token-level context, while $s_T$ compactly summarizes
the net effect of the refinement process. The final hidden state used for
decoding (denoted $\tilde{h}_T$, distinct from the rollout state $h_T$) is
thus:
\[
  \tilde{h}_T = f_{\mathrm{decode}}(s_T, h_0),
  \qquad
  \hat{y} \sim p_{\theta}(y \mid x, \tilde{h}_T).
\]
$f_{\mathrm{decode}}$ is implemented as a gated interpolation between $h_0$
and a projection of $s_T$: writing $\mathrm{proj}(s_T)\in\mathbb{R}^{D}$ for a
two-layer MLP projection of $s_T$ and $g(s_T)\in[0,1]^{D}$ for a
sigmoid-gated network conditioned on $s_T$, the final representation is
\[
  \tilde{h}_T = \mathrm{LayerNorm}\Big((1-g(s_T))\odot h_0 + g(s_T)\odot \mathrm{proj}(s_T)\Big),
\]
where $g(s_T)$ is broadcast identically across all token positions.

\subsection{Reinforcement-Learning Controller Optimization}
\label{sec:rl}
The refinement controllers are trained to make latent updates that improve the
likelihood of the target output while avoiding unnecessary computation. During
training, the target sequence is available, so we can evaluate whether each
refinement step makes the current representation more predictive of the correct
answer. This provides a natural reward signal for optimizing both the depth
controller and the action controller.
\paragraph{Adaptive depth.}
We introduce a depth controller $\pi_d$ that selects the number of refinement
steps from the initial reasoning state:
\[
  T \sim \pi_d(T \mid s_0).
\]
Concretely, $\pi_d$ is parameterized as a categorical distribution over
$\{0,1,\ldots,T_{\max}\}$ conditioned on $s_0$. This allows the model to
allocate different amounts of latent computation to different inputs. Easier
examples may require only a few refinement steps, whereas more difficult
examples can receive additional updates.
\paragraph{Step-wise reward.}
For any hidden state $h_t$, the per-step improvement and reward are:
\[
\begin{aligned}
\Delta_t
&=
\log p_\theta(y^* \mid x, h_{t+1})
-
\log p_\theta(y^* \mid x, h_t), \\
r_t
&= \Delta_t - c_d,
\end{aligned}
\]
where $c_d > 0$ is a fixed per-step computation cost. The improvement term
encourages updates that increase the likelihood of the target sequence, while
the computation cost discourages unnecessary refinement. Note that $r_t$ is
computed by hypothetically decoding directly from $h_{t+1}$, whereas the final
prediction instead decodes from $f_{\mathrm{decode}}(s_T, h_0)$. We use this
per-step likelihood gain as a proxy reward because it provides dense credit
assignment for each latent action, rather than only a sparse final-output
signal. Intuitively, updates that make $h_{t+1}$ more predictive of $y^*$ are
encouraged because they are likely to contribute useful information to
$s_{t+1}$ through $g(\cdot)$. Empirically, Figure~4 shows that likelihood gains
along the $h_t$ trajectory track improvements in the final decoded
representation.
\paragraph{Shaped return and RL objective.}
Here, $\mathrm{KL}$ denotes the token-averaged forward KL divergence between
the full output distributions of the policy model (conditioned on the refined
representation) and a frozen reference copy of the same pretrained model
(conditioned on the unrefined input):
\[
\begin{aligned}
\mathrm{KL}
&=
\frac{1}{M}\sum_{i=1}^{M}\sum_{v\in\mathcal{V}}
p_\theta(v\mid x,\tilde{h}_T)_i \\
&\quad \times
\Big[
\log p_\theta(v\mid x,\tilde{h}_T)_i
-
\log p_{\mathrm{ref}}(v\mid x)_i
\Big].
\end{aligned}
\]
where the sum over $i$ runs over valid target positions, $\mathcal{V}$ is the
model's vocabulary, and $p_{\mathrm{ref}}$ denotes the reference model's
output distribution obtained from a single forward pass without any latent
refinement. This term anchors the refined model to the original pretrained
distribution and discourages the refinement process from drifting into
degenerate output distributions.
The shaped step return is:
\[
  R_t
  =
  \frac{r_t}{t+1}
  +
  \frac{1}{T}\!\left(-\beta_{\mathrm{KL}}\,\mathrm{KL} + \lambda H\right),
\]
and the policy-gradient RL loss is:
\[
\mathcal{L}_{\mathrm{RL}}
  =
  -\mathbb{E}\!\left[
    A \cdot
    \left(
      \log \pi_d(T \mid s_0)
      +
      \sum_{t=0}^{T-1} \log \pi_a(a_t \mid s_t)
    \right)
  \right],
\]
where $A = R - \mathbb{E}[R]$ is the advantage estimate. This objective
reinforces depth and action decisions that lead to positive refinement returns,
while the entropy term encourages sufficient exploration during training.
In practice, we estimate $\mathbb{E}[R \mid s_0]$ with a learned value
critic $V(s_0)$ rather than a batch-level average, following standard
actor-critic variance reduction; the critic is trained jointly via an
auxiliary MSE loss.

\subsection{Joint Training Objective}
\label{sec:training}

After latent refinement, the final representation $\tilde{h}_T$ is used for standard
autoregressive prediction. We optimize the language modeling objective over the
target sequence $y^*$:
\[
  \mathcal{L}_{\mathrm{LM}}
  =
  -\frac{1}{M}
  \sum_{t=1}^{M}
  \log p_\theta\!\left(y^*_t \mid y^*_{<t}, x, \tilde{h}_T\right).
\]
This term ensures that the refined representation remains useful for generating
the correct output sequence.

The full training objective combines the language modeling loss with the
reinforcement-learning loss for the refinement controllers:
\[
  \mathcal{L}_{\mathrm{total}}
  =
  \mathcal{L}_{\mathrm{LM}}
  +
  \alpha_{\mathrm{RL}}\,\mathcal{L}_{\mathrm{RL}},
\]
where $\alpha_{\mathrm{RL}}$ balances language modeling fidelity against the
quality of learned refinement policies. In this objective, the language modeling
loss trains the model to produce the target output from the refined hidden
state, while the RL loss trains the controllers to select refinement depths and
actions that improve the latent representation before decoding.

\begin{table*}[t]
\centering
\caption{Performance comparison on four reasoning datasets under 0-shot and 5-shot settings. 
Ours denotes ReLAR with Gemma-2B. Bold indicates the best result in each column.}
\label{tab:main-results}
\resizebox{\textwidth}{!}{%
\begin{tabular}{lcc cc cc cc}
\toprule
\multirow{2}{*}{\textbf{Models}} &
\multicolumn{2}{c}{\textbf{PubMedQA (Acc./F1)}} &
\multicolumn{2}{c}{\textbf{GSM8K (Acc./pass@5)}} &
\multicolumn{2}{c}{\textbf{GSM-Hard (Acc./pass@5)}} &
\multicolumn{2}{c}{\textbf{HotpotQA (Acc./F1)}} \\
\cmidrule(lr){2-3}\cmidrule(lr){4-5}\cmidrule(lr){6-7}\cmidrule(lr){8-9}
& \textbf{0-shot} & \textbf{5-shot}
& \textbf{0-shot} & \textbf{5-shot}
& \textbf{0-shot} & \textbf{5-shot}
& \textbf{0-shot} & \textbf{5-shot} \\
\midrule
LLaMA-2-7B
  & 53.47/46.83 & 58.34/51.67
  & 23.82/31.47 & 30.56/48.73
  & 22.34/27.83 & 29.47/38.57
  & 27.34/38.47 & 39.82/52.63 \\
Mistral-7B-v0.3
  & 59.83/52.47 & 64.57/57.83
  & 40.63/67.84 & 57.84/79.36
  & 24.83/31.47 & 38.64/46.28
  & 35.63/48.74 & 47.82/61.35 \\
Falcon-7B
  & 44.83/37.47 & 49.64/42.83
  & 17.84/24.47 & 25.63/36.82
  & 14.47/19.83 & 21.28/27.64
  & 22.47/31.83 & 30.84/41.64 \\
Gemma-7B
  & 56.84/49.73 & 61.47/54.28
  & 47.63/68.84 & 56.28/76.43
  & 22.47/29.63 & 31.84/39.47
  & 31.82/43.47 & 42.64/56.83 \\
Mistral-7B-Instruct
  & 63.28/55.84 & 67.47/60.83
  & 53.47/74.83 & 63.82/82.47
  & 27.83/34.47 & 36.28/44.83
  & 41.47/55.28 & 50.83/64.47 \\
Llama-3-8B-Instruct
  & 68.47/62.83 & 71.83/66.47
  & \textbf{75.82/83.47} & \textbf{80.64/88.23}
  & 34.47/42.83 & 43.28/51.84
  & 48.83/63.47 & 56.47/71.83 \\
Qwen2.5-7B
  & 58.92/51.47 & 63.84/56.23
  & 73.47/81.83 & 79.28/87.64
  & 39.83/\textbf{47.28} & 47.64/54.83
  & 44.83/58.47 & 52.64/66.83 \\
Qwen2.5-Med-7B
  & 60.12/52.48 & 65.09/57.34
  & 64.41/73.85 & 71.35/80.63
  & 36.47/43.82 & 42.82/50.03
  & 41.81/55.13 & 50.18/64.15 \\
Med42-Mistral-7B
  & 69.14/60.53 & 71.38/63.24
  & 46.53/72.84 & 54.82/81.47
  & 30.63/38.47 & 38.84/47.23
  & 32.47/45.83 & 39.64/53.47 \\
Med42-Llama3-8B
  & 70.65/70.74 & 73.87/72.25
  & 71.89/80.14 & 77.06/85.80
  & 35.62/42.93 & 43.49/51.63
  & 46.23/61.59 & 54.01/69.82 \\
MedGemma-4B
  & 72.45/68.52 & 74.19/70.71
  & 61.18/70.32 & 66.71/75.29
  & 26.03/33.20 & 34.63/42.08
  & 39.48/53.29 & 46.91/60.05 \\
\midrule
\textbf{Ours}
  & \textbf{77.67/72.54} & \textbf{79.23/74.17}
  & 68.45/78.23 & 71.28/84.20
  & \textbf{41.06}/45.58  & \textbf{48.57/52.12}
  & \textbf{57.50/75.23} & \textbf{59.64/76.15} \\
\bottomrule
\end{tabular}%
}
\end{table*}

\begin{table}[!htbp]
\centering
\caption{Open-ended generation performance across general-domain generation tasks.}
\label{tab:open-ended-generation}
\resizebox{\columnwidth}{!}{%
\begin{tabular}{lcc cc}
\hline
\multirow{2}{*}{\textbf{Models}} &
\multicolumn{2}{c}{\textbf{CommonGen}} &
\multicolumn{2}{c}{\textbf{WritingPrompts}} \\
\cline{2-5}
& \textbf{BERTScore} & \textbf{ROUGE-L}
& \textbf{BERTScore} & \textbf{ROUGE-L} \\
\hline
LLaMA-2-7B          & 0.884 & 21.37 & 0.847 &  5.12 \\
Mistral-7B-v0.3     & 0.903 & 27.84 & 0.858 &  6.83 \\
Gemma-7B            & 0.896 & 24.63 & 0.853 &  6.17 \\
Qwen2.5-7B          & 0.912 & 31.42 & 0.864 &  7.94 \\
Mistral-7B-Instruct & 0.921 & 35.67 & 0.871 &  9.28 \\
Llama-3-8B-Instruct & 0.918 & 33.94 & 0.869 &  8.74 \\
Falcon-7B           & 0.879 & 19.28 & 0.843 &  4.63 \\
\hline
\textbf{Ours}
  & \textbf{0.934} & \textbf{38.92}
  & \textbf{0.878} & \textbf{11.47} \\
\hline
\end{tabular}%
}
\end{table}

\begin{table*}[t]
\centering
\caption{Ablation study on refinement strategies across four reasoning datasets. All variants use Gemma-2B. Bold indicates the best result in each column.}
\label{tab:ablation}
\resizebox{\textwidth}{!}{%
\begin{tabular}{l cc cc cc cc}
\toprule
\multirow{2}{*}{\textbf{Ablation Setting}} &
\multicolumn{2}{c}{\textbf{PubMedQA}} &
\multicolumn{2}{c}{\textbf{GSM8K}} &
\multicolumn{2}{c}{\textbf{GSM-Hard}} &
\multicolumn{2}{c}{\textbf{HotpotQA}} \\
\cmidrule(lr){2-3}\cmidrule(lr){4-5}\cmidrule(lr){6-7}\cmidrule(lr){8-9}
& Acc. & F1 & Acc. & pass@5 & Acc. & pass@5 & Acc. & F1 \\
\midrule
No Refinement (SFT only)
  & 58.10 & 33.54 & 48.52 & 63.41 & 29.14 & 32.37 & 34.82 & 36.15 \\
Static Refinement (fixed $d$, no direction)
  & 73.01 & 60.84 & 63.84 & 74.18 & 36.52 & 40.26 & 52.83 & 63.45 \\
w/ Adaptive Depth (no direction)
  & 76.72 & 57.85 & 66.72 & 77.14 & \textbf{42.28} & \textbf{46.83} & 56.61 & 60.57 \\
w/ Adaptive Direction (no depth, single-step)
  & 68.90 & 57.12 & 60.21 & 71.32 & 33.85 & 37.51 & 48.71 & 59.82 \\
\midrule
\textbf{Ours} (Adaptive Depth + Direction)
  & \textbf{77.67} & \textbf{72.54}
  & \textbf{68.45} & \textbf{78.23}
  & 41.06 & 45.58
  & \textbf{57.50} & \textbf{75.23} \\
\bottomrule
\end{tabular}
}
\end{table*}





\begin{table}[t]
\centering
\caption{Accuracy--latency comparison on PubMedQA using Gemma-2B. 
ICL denotes 5-shot in-context learning, SC-CoT uses $n=5$ sampled reasoning paths, and Time/Ours reports inference time relative to ReLAR.}
\label{tab:latency}
\begin{tabular}{lcccc}
\toprule
\textbf{Method} & \textbf{Acc.} & \textbf{F1} & \textbf{Time (s)} & \textbf{Time/Ours} \\
\midrule
SFT only   & 58.10 & 33.54 & 0.12 & 0.9$\times$ \\
ICL        & 73.47 & 70.46 & 0.31 & 2.2$\times$ \\
CoT        & 64.68 & 54.33 & 9.09 & 64.9$\times$ \\
SC-CoT     & 72.83 & 65.31 & 16.36 & 116.9$\times$ \\
\textbf{Ours} & \textbf{77.67} & \textbf{72.54} & 0.14 & 1.0$\times$ \\
\bottomrule
\end{tabular}
\end{table}

\begin{table}[t]
\centering
\caption{Comparison between SFT and latent refinement on PubMedQA.}
\label{tab:sft-comparison}
\renewcommand{\arraystretch}{1.12}
\begin{tabular*}{\columnwidth}{@{\extracolsep{\fill}}lcc}
\hline
\textbf{Model Variant} & \textbf{Acc.} & \textbf{F1} \\
\hline
LLaMA-1.1B + SFT        & 52.02 & 48.68 \\
LLaMA-1.1B + Ours       & 65.22 & 58.50 \\
\hline
Gemma-2B + SFT          & 58.10 & 33.54 \\
Gemma-2B + Ours         & 77.67 & 72.54 \\
\hline
Qwen-3B + SFT           & 60.45 & 51.12 \\
Qwen-3B + Ours          & 74.88 & 69.84 \\
\hline
\end{tabular*}
\end{table}

\section{Experiments}
\label{sec:experiments}

\subsection{Experimental Setting}

\textbf{Datasets.}
We evaluate on six benchmarks that cover complementary aspects of reasoning
and generation. PubMedQA~\citep{jin2019pubmedqa} is used to evaluate
biomedical question answering, where reliable reasoning is important because
answers must be selected from yes/no/maybe labels grounded in scientific
abstracts. We report accuracy to measure final-answer correctness and macro-F1
to account for class imbalance across answer types. GSM8K~\citep{cobbe2021training}
and GSM-Hard are used to evaluate multi-step arithmetic reasoning, with
GSM-Hard containing more challenging numerical variants. We report accuracy for
top-1 correctness and pass@5 to measure whether the correct solution can be
recovered among multiple sampled attempts. HotpotQA~\citep{yang2018hotpotqa}
is used to evaluate multi-hop question answering across multiple documents,
testing whether the model can integrate evidence from different contexts. We
report accuracy and F1 to capture both exact answer correctness and token-level
overlap with reference answers. For open-ended generation, we evaluate on
CommonGen~\citep{lin2020commongen} and
WritingPrompts~\citep{fan2018hierarchical}, which test constrained
commonsense generation and long-form creative generation, respectively. We use
BERTScore to measure semantic similarity and ROUGE-L to measure lexical and
sequence-level overlap with reference outputs.\\

\textbf{Implementation details.}
We use Gemma-2B as the primary backbone for ReLAR. Unless otherwise specified,
all main experiments are conducted with Gemma-2B. To examine whether the proposed
latent refinement mechanism generalizes across different lightweight backbones,
we additionally implement ReLAR on LLaMA-1.1B and Qwen-3B, and report controlled
same-backbone comparisons against their corresponding SFT-only variants. All models are trained with a learning rate of $1 \times 10^{-6}$, a maximum sequence length of 512, and a maximum refinement depth of $T_{\max}=3$. Training is performed on a single NVIDIA A100 80GB GPU.

\subsection{Baselines}

We compare ReLAR against both general-purpose and domain-specific medical
LLMs. General-purpose baselines include LLaMA-2-7B, Mistral-7B-v0.3,
Falcon-7B, Gemma-7B, Mistral-7B-Instruct, Llama-3-8B-Instruct, and
Qwen2.5-7B. Medical baselines include Qwen2.5-Med-7B, Med42-Mistral-7B,
Med42-Llama3-8B~\citep{christophe2024med42}, and
MedGemma-4B~\citep{sellergren2025medgemma}. We also include SFT-only,
static-refinement, adaptive-depth-only, and adaptive-direction-only variants
as controlled ablations to isolate the contribution of each ReLAR component.

\subsection{Evaluation Protocol}

For PubMedQA, we follow the standard three-way classification setting
and report accuracy and macro-F1. For GSM8K and GSM-Hard, we report
accuracy and pass@5 to evaluate both direct correctness and sampled
solution quality. For HotpotQA, we report accuracy and F1 to measure
multi-hop answer correctness. For open-ended generation, we compute
BERTScore~\citep{zhang2020bertscore} and ROUGE-L~\citep{lin2004rouge}
against reference outputs to assess semantic similarity and sequence-level
overlap, respectively. All reported numbers are averaged over three
random seeds; we use the same decoding temperature and maximum generation
length for all models to ensure comparability.

\subsection{Main Results}

Table~\ref{tab:main-results} summarizes the results on four reasoning
benchmarks under both 0-shot and 5-shot settings. On PubMedQA, ReLAR achieves
the best performance among all compared models, obtaining 77.67\% accuracy and
72.54 macro-F1 in the 0-shot setting, and 79.23\% accuracy and 74.17 macro-F1
in the 5-shot setting. Compared with the strongest medical baseline,
MedGemma-4B, ReLAR improves 0-shot accuracy by 5.22 percentage points and
5-shot accuracy by 5.04 percentage points. These gains suggest that latent
hidden-state refinement is particularly effective for biomedical question
answering, where the model must integrate evidence from scientific contexts and
produce reliable yes/no/maybe decisions.

On mathematical reasoning tasks, ReLAR shows competitive performance on GSM8K
and stronger gains on the more challenging GSM-Hard benchmark. On GSM8K, ReLAR
achieves 68.45\% accuracy and 78.23 pass@5 in the 0-shot setting, and 71.28\%
accuracy and 84.20 pass@5 in the 5-shot setting. Although larger instruction
models such as Llama-3-8B-Instruct obtain higher GSM8K accuracy, ReLAR remains
competitive while using lightweight backbones and an implicit latent refinement
mechanism rather than explicit reasoning traces. On GSM-Hard, ReLAR achieves the
highest accuracy under both shot settings, with 41.06\% accuracy in 0-shot and
48.57\% accuracy in 5-shot. This indicates that adaptive latent refinement is
especially useful when numerical reasoning becomes more difficult and standard
solution patterns are less reliable.

On HotpotQA, ReLAR achieves the strongest overall performance, reaching 57.50\%
accuracy and 75.23 F1 in the 0-shot setting, and 59.64\% accuracy and 76.15 F1
in the 5-shot setting. Compared with Llama-3-8B-Instruct, ReLAR improves 0-shot
accuracy by 8.67 percentage points and 0-shot F1 by 11.76 points. These gains
show that iterative latent refinement is beneficial for multi-hop reasoning,
where the model must combine evidence across multiple pieces of context before
producing the final answer.

Table~\ref{tab:open-ended-generation} reports open-ended generation results on
CommonGen and WritingPrompts. ReLAR achieves the highest score on all reported
metrics, with a BERTScore of 0.934 and ROUGE-L of 38.92 on CommonGen, and a
BERTScore of 0.878 and ROUGE-L of 11.47 on WritingPrompts. The improvements are
consistent across both semantic similarity and lexical overlap metrics,
suggesting that the proposed latent refinement mechanism improves not only
reasoning accuracy but also the quality of free-form generation.

\begin{figure}[t]
    \centering
    \includegraphics[width=\columnwidth]{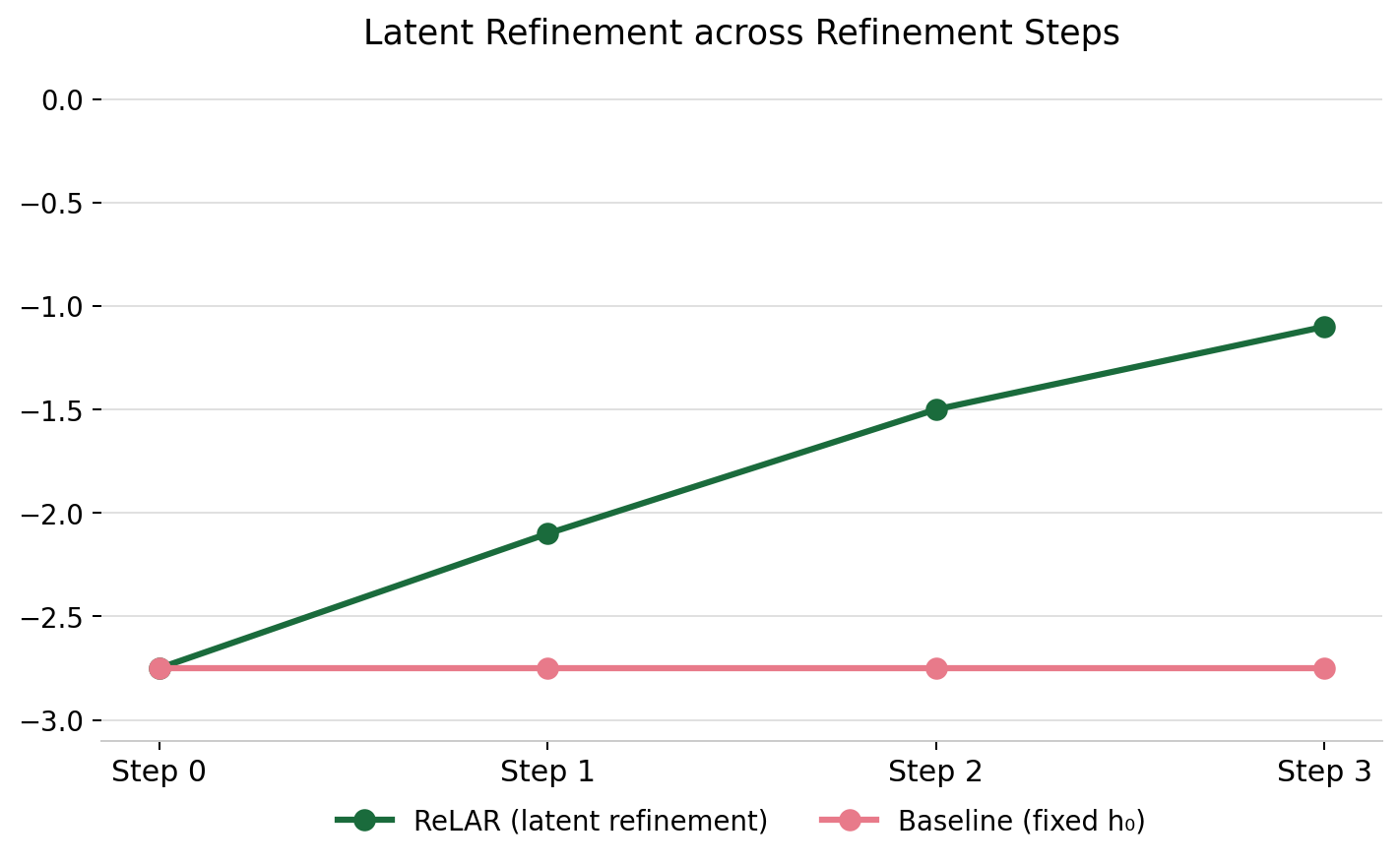}
    \caption{Latent refinement across refinement steps. ReLAR 
    progressively improves while the baseline remains flat.}
    \label{fig:log_likelihood}
\end{figure}

\subsection{Ablation Studies}

\textbf{Refinement strategy.}
Table~\ref{tab:ablation} evaluates the contribution of different refinement
strategies. Removing latent refinement leads to the weakest performance across
all datasets, with 58.10 accuracy and 33.54 F1 on PubMedQA, 48.52 accuracy on
GSM8K, 29.14 accuracy on GSM-Hard, and 34.82 accuracy on HotpotQA. Introducing
static refinement with a fixed depth substantially improves over the SFT-only
model, increasing PubMedQA accuracy from 58.10 to 73.01 and HotpotQA accuracy
from 34.82 to 52.83. These gains show that refining hidden representations
before decoding is an effective mechanism for improving reasoning performance.
However, static refinement remains less effective than adaptive refinement,
indicating that input-dependent control is important for allocating refinement
computation to examples with different levels of reasoning complexity.

Adaptive depth and adaptive direction further improve this refinement process.
Adaptive depth enables the model to adjust the amount of latent computation
based on input difficulty, yielding strong gains on challenging reasoning
benchmarks such as PubMedQA and GSM-Hard. Adaptive direction provides an
additional mechanism for controlling how the hidden representation is updated,
leading to consistent improvements over the SFT-only baseline. When combined,
adaptive depth and adaptive direction produce the strongest overall performance
across the evaluated tasks, achieving 77.67 accuracy and 72.54 F1 on PubMedQA,
68.45 accuracy and 78.23 pass@5 on GSM8K, 41.06 accuracy and 45.58 pass@5 on
GSM-Hard, and 57.50 accuracy and 75.23 F1 on HotpotQA. These results demonstrate
that ReLAR benefits from both deciding how much refinement to perform and
controlling the direction of each latent update, leading to robust gains across
medical, mathematical, and multi-hop reasoning tasks.

\textbf{Backbone comparison.}
Table~\ref{tab:sft-comparison} shows that ReLAR consistently improves over
SFT-only training across all three lightweight backbones on PubMedQA. For
LLaMA-1.1B, ReLAR improves accuracy from 52.02 to 65.22 and F1 from 48.68 to
58.50. For Gemma-2B, the gain is larger, improving accuracy from 58.10 to 77.67
and F1 from 33.54 to 72.54. For Qwen-3B, ReLAR improves accuracy from 60.45 to
74.88 and F1 from 51.12 to 69.84. These consistent gains indicate that latent
refinement provides benefits beyond standard supervised fine-tuning and is not
limited to a single backbone architecture.

\textbf{Efficiency comparison.}
We compare ReLAR with explicit reasoning baselines to assess whether latent
refinement can improve reasoning accuracy without incurring the generation cost
of textual reasoning traces. On PubMedQA with the Gemma-2B backbone,
Table~\ref{tab:latency} shows that ReLAR achieves the best accuracy and F1, with
77.67 accuracy and 72.54 F1, while requiring only 0.14 seconds per example. In
contrast, Chain-of-Thought (CoT) and Self-Consistency Chain-of-Thought
(SC-CoT) require 9.09 and 16.36 seconds, respectively, corresponding to
64.9$\times$ and 116.9$\times$ higher inference time than ReLAR. In-Context
Learning (ICL), evaluated in a 5-shot setting, is faster than CoT-based methods
but remains 2.2$\times$ slower than ReLAR and achieves lower accuracy and F1.
The CoT and ICL prompt templates are provided in Section~3 of the supplementary
material.

\textbf{Refinement dynamics.}
We further analyze the refinement dynamics to verify whether the iterative
latent updates progressively improve the model's internal representation before
decoding. We use the target log-likelihood trajectory across refinement stages
as a proxy for reasoning stability, since stable refinement should progressively
make the internal representation more predictive of the target output rather
than causing degraded or oscillatory likelihood. Specifically, this experiment
tracks the target log-likelihood at each possible refinement stage up to the
maximum refinement depth $T_{\max}=3$ and compares ReLAR with a
fixed-hidden-state baseline that does not perform adaptive latent updates. In
Figure~\ref{fig:log_likelihood}, the x-axis denotes the refinement stage: Step
0 corresponds to the initial hidden state before refinement, and Steps 1--3
correspond to the hidden states after one, two, and three latent refinement
updates, respectively. Since ReLAR uses adaptive depth, not all inputs
necessarily execute all three updates; therefore, the plot summarizes the
average likelihood trajectory over the evaluated refinement stages up to the
maximum depth $T_{\max}=3$. The y-axis reports the target log-likelihood, with
higher values indicating that the hidden representation better supports the
target output. ReLAR steadily increases the target likelihood across refinement
stages, whereas the fixed-hidden-state baseline remains nearly flat. This
indicates that the learned refinement module is not merely adding a static
transformation, but progressively adjusts the latent representation toward a
more predictive state.

\section{Conclusion}
We presented ReLAR, a reinforcement-guided latent 
refinement framework that enables controllable 
multi-step reasoning entirely within the hidden-state 
space of a pretrained language model. Rather than 
relying on explicit chain-of-thought generation, 
ReLAR iteratively refines internal representations 
prior to decoding, guided by learned depth and action 
controllers trained with a policy-gradient objective. 
Experiments across medical, mathematical, multi-hop, 
and open-ended generation benchmarks demonstrate that 
ReLAR consistently improves accuracy and generation 
quality over strong general-purpose and medical LLM 
baselines, while achieving substantially lower 
inference overhead than explicit reasoning approaches. 
Ablation studies further confirm that both adaptive 
depth and adaptive direction contribute to the overall 
performance, and that reinforcement-guided refinement 
provides complementary benefits beyond standard 
supervised fine-tuning.

\section*{Limitations}
Despite promising results, ReLAR has several 
limitations. First, our backbone models (LLaMA-1.1B, 
Gemma-2B, Qwen-3B) are smaller than the 7B baselines 
in our comparison, which may limit the direct 
comparability of results. Second, the reinforcement 
learning training requires per-step likelihood 
evaluation against ground-truth labels, making the 
framework dependent on supervised signal and 
potentially less applicable to purely unsupervised 
settings. Third, while latent refinement reduces 
inference overhead compared to chain-of-thought 
approaches, the iterative hidden-state updates still 
introduce additional parameters and training 
complexity relative to standard fine-tuning. Finally, 
the internal refinement process operates entirely in 
latent space and is not directly interpretable, which 
may limit applicability in settings where reasoning 
transparency is required, such as high-stakes clinical 
decision support.


\

\appendix
\onecolumn




\section{Algorithmic Details}
\label{apd:algorithm}

Algorithm~\ref{alg:relar_training} summarizes the training procedure of ReLAR.
The procedure first encodes the input into an initial hidden representation,
constructs a compact reasoning state, and samples an input-dependent refinement
depth using the depth controller. During training, the action controller
iteratively updates the hidden representation in latent space. The step-wise
reward is computed from the likelihood improvement of the target sequence after
each latent update, and the depth and action controllers are optimized with a
policy-gradient objective.

\begin{algorithm}[t]
\caption{Training procedure of ReLAR with RL-guided latent refinement.}
\label{alg:relar_training}
\small
\begin{minipage}{0.97\linewidth}
\textbf{Input:} input \(x\), target \(y^*\), LM \(p_\theta\), max depth \(T_{\max}\).\\
\textbf{Initialize:} obtain \(h_0\) from \(p_\theta\), set
\(s_0=f_{\mathrm{extract}}(h_0)\), sample
\(T\sim\pi_d(T\mid s_0)\) with \(T\leq T_{\max}\), and set \(R\leftarrow 0\).
\medskip
\textbf{Latent refinement.}
For \(t=0,\ldots,T-1\):
\[
\begin{aligned}
a_t=(\gamma_t,\beta_t,v_t) &\sim \pi_a(a_t\mid s_t), \qquad
v_t \leftarrow v_t/\|v_t\|_2,\\
\alpha_t &= f_\alpha(\gamma_t,\beta_t),\\
h_{t+1} &= h_t+\alpha_t v_t,\\
s_{t+1} &= g(s_t,h_{t+1}),\\
\Delta_t &=
\log p_\theta(y^*\mid x,h_{t+1})
-
\log p_\theta(y^*\mid x,h_t),\\
r_t &= \Delta_t-c_d,\\
R_t &=
\frac{r_t}{t+1}
+
\frac{1}{T}
\left(-\beta_{\mathrm{KL}}\mathrm{KL}+\lambda H\right),\\
R &\leftarrow R+R_t .
\end{aligned}
\]
\textbf{Optimization.}
After refinement, decode the final representation
\(\tilde{h}_T=f_{\mathrm{decode}}(s_T,h_0)\) and compute
\[
\mathcal{L}_{\mathrm{LM}}
=
-\log p_\theta(y^*\mid x,\tilde{h}_T),
\qquad
A=R-\mathbb{E}[R],
\]
\[
\mathcal{L}_{\mathrm{RL}}
=
-A
\left(
\log \pi_d(T\mid s_0)
+
\sum_{t=0}^{T-1}\log \pi_a(a_t\mid s_t)
\right),
\]
\[
\mathcal{L}_{\mathrm{total}}
=
\mathcal{L}_{\mathrm{LM}}
+
\alpha_{\mathrm{RL}}\mathcal{L}_{\mathrm{RL}}.
\]
\end{minipage}
\end{algorithm}

At inference time, the same latent refinement procedure is applied without
reward computation because the target sequence is unavailable. The trained depth
controller selects the refinement depth, while the trained action controller
produces the latent update at each step before final decoding.

\section{Prompt Templates}
\label{apd:prompt_templates}

\subsection{Chain-of-Thought Prompt}
\label{apd:cot_prompt}

The Chain-of-Thought (CoT) prompt asks the model to explicitly follow four
labeled reasoning steps before producing a categorical answer. The model is instructed not to skip, merge, or reorder the reasoning steps. The final output is constrained to the format \texttt{Answer: yes} or \texttt{Answer: no} or \texttt{Answer: maybe}.

\begin{tcolorbox}[
    colback=gray!5,
    colframe=gray!70,
    title=CoT Prompt Template,
    breakable,
    fonttitle=\bfseries
]
\begin{verbatim}
You are a biomedical question-answering system.

You must reason using the following 4 labeled steps.
Do not skip, merge, or reorder the steps.
Each step must be clearly labeled and on a separate line.

After reasoning, provide a final answer in this exact format:
"Answer: yes" or "Answer: no" or "Answer: maybe"

Example 1:
Question: Do mitochondria play a role in remodelling lace plant leaves
during programmed cell death?
1. Question Decomposition: Determine if mitochondria are involved in a
specific form of cell death in lace plant leaves.
2. Background Knowledge: Mitochondria regulate apoptosis in both animal
and plant cells, and are active during plant programmed cell death.
3. Logical Connection: Since mitochondria are involved in PCD in plants,
and lace plants undergo PCD during remodelling, mitochondria likely
participate.
4. Final Justification: Based on known plant cell biology, mitochondria
are involved in remodelling lace plant leaves during PCD.
Answer: yes

Now, answer the following:

Question: {question}
1. Question Decomposition:
\end{verbatim}
\end{tcolorbox}

\subsection{In-Context Learning Prompt}
\label{apd:icl_prompt}

The In-Context Learning (ICL) prompt uses a 5-shot setting, where five biomedical question-answer examples are provided before the target question. The model is instructed to answer strictly with either \texttt{yes} or \texttt{no} or \texttt{maybe}. When external evidence is available, it is inserted as the context before the target question.

\begin{tcolorbox}[
    colback=gray!5,
    colframe=gray!70,
    title=ICL Prompt Template,
    breakable,
    fonttitle=\bfseries
]
\begin{verbatim}
You are a biomedical QA system.
Answer strictly with 'yes' or 'no' or 'maybe'.

Example 1:
Question: Is aspirin an antiplatelet drug?
Answer: yes

Example 2:
Question: Is insulin a steroid hormone?
Answer: no

Example 3:
Question: Does smoking increase the risk of lung cancer?
Answer: yes

Example 4:
Question: Are antibiotics used to treat bacterial infections?
Answer: yes

Example 5:
Question: Is hemoglobin primarily responsible for blood clotting?
Answer: no

Now answer:

Context: {context}
Question: {question}
Answer:
\end{verbatim}
\end{tcolorbox}

\section{Reproducibility Details}
This section provides additional details to facilitate reproduction of the reported
experiments. The main experiments use Gemma-2B as the backbone model, a learning rate of
$1 \times 10^{-6}$, a maximum sequence length of 512, and a maximum refinement depth of
$T_{\max}=3$. Training is performed on a single NVIDIA A100 80GB GPU.

\subsection{Training Configuration}
Unless otherwise specified, all ReLAR models are trained with the same optimization
configuration. We use AdamW with a base learning rate of $1 \times 10^{-6}$, scaled
per module (0.1$\times$ for the base LM; 2--3$\times$ for the newly introduced
state extractor, depth controller, action controller, refiner, decode bridge, and
value critic), a batch size of 2, weight decay of 0.01, and a gradient clipping
threshold of 1.0. The maximum input length is set to 512 tokens, and the maximum
latent refinement depth is $T_{\max}=3$. The reinforcement-learning loss weight is
$\alpha_{RL}=0.15$, the per-step computation cost is $c_d=0.02$, the KL
regularization coefficient is $\beta_{KL}=0.1$, and the entropy coefficient is
$\lambda=0.01$. The reasoning state and action dimensions are $d_s=256$ and
$d_v=64$, respectively. All random seeds (data shuffling, module initialization)
are fixed to 42.

\subsection{Value Critic Details}
The advantage baseline described in Section~3.4 of the main paper is estimated
with a learned value critic $V(s_0)$, implemented as a lightweight multi-layer
perceptron that takes the initial reasoning state $s_0 \in \mathbb{R}^{d_s}$ as
input and outputs a scalar value estimate. The critic is trained jointly with
the depth and action controllers via an auxiliary mean-squared-error loss,
\[
  \mathcal{L}_{\text{critic}} = \big(V(s_0) - R\big)^2,
\]
where $R$ is the realized shaped return for the corresponding trajectory. This
loss is weighted by a coefficient of 0.2 in the total training objective. Using
$V(s_0)$ in place of a batch-level average $\mathbb{E}[R]$ allows the baseline
to condition on the difficulty of the input, following standard actor-critic
variance reduction; the critic itself receives no gradient from the policy loss
and is optimized solely to minimize $\mathcal{L}_{\text{critic}}$.

\subsection{Refinement Direction Projection}
The action-controller output $v_t \in \mathbb{R}^{d_v}$ is mapped back into the
hidden-state space via a learned linear projection $v_t \mapsto \mathbb{R}^{D}$,
re-normalized to unit norm, and broadcast identically across all $L$ token
positions before being added to $h_t \in \mathbb{R}^{L\times D}$. This
projection step is omitted from the simplified update $h_{t+1}=h_t+\alpha_t v_t$
in the main paper for brevity.

\subsection{Answer Extraction}
For PubMedQA, model outputs are normalized by extracting the first valid answer label among
``yes'', ``no'', and ``maybe''. For GSM8K and GSM-Hard, we extract the final numerical
answer by locating all numeric substrings in the generated text and taking the last
occurrence as the predicted value. For HotpotQA, we normalize predictions by
lowercasing, removing punctuation and articles, and applying the standard token-level F1
evaluation script. Invalid or empty generations are treated as incorrect.

\subsection{Dataset Processing}
We use the official dataset splits when available. For GSM-Hard, we use a 90/10
train/validation split (seed 42) over the \texttt{reasoning-machines/gsm-hard} release.
For PubMedQA, the scientific abstract is used as context and concatenated with the question.
For HotpotQA, the provided supporting contexts are concatenated before the question. For
GSM8K, the input consists of the problem statement. For CommonGen and WritingPrompts, the
input follows the standard task format. Prompts exceeding the length budget are truncated
from the right (via tokenizer truncation) so that the full target sequence is always
preserved within the 512-token limit.

\subsection{Baseline Evaluation}
All baseline models are evaluated from official checkpoints without additional fine-tuning.
To ensure fair comparison, baselines and ReLAR use the same prompt templates, input
formatting, decoding configuration, and answer extraction scripts. The SFT-only baseline is
trained on the same data as ReLAR but removes the latent refinement module and the
reinforcement-guided depth and action controllers.

\subsection{Random Seeds and Released Artifacts}
We fix random seeds for data shuffling, newly initialized modules, and stochastic decoding
(seed = 42). Upon acceptance, we will release the training scripts, evaluation scripts,
preprocessing code, prompt templates, configuration files, answer extraction scripts, and
trained controller checkpoints used to produce the GSM-Hard results reported in this paper;
scripts for the remaining benchmarks will be released concurrently.

\section{Notation Summary}
\label{sec:notation}
Table~\ref{tab:notation} summarizes the main symbols used throughout the
paper.

\begin{table}[h]
\centering
\begin{tabular}{ll}
\toprule
Symbol & Description \\
\midrule
$x$ & Input token sequence \\
$y^*$ & Target output sequence \\
$h_0$ & Initial hidden state from the base LM, $h_0\in\mathbb{R}^{L\times D}$ \\
$h_t$ & Hidden state at refinement step $t$ (auxiliary training rollout) \\
$\tilde{h}_T$ & Final hidden state used for decoding, $\tilde{h}_T=f_{\mathrm{decode}}(s_T,h_0)$ \\
$s_t$ & Reasoning state at step $t$, $s_t\in\mathbb{R}^{d_s}$ \\
$T$ & Number of refinement steps, sampled from $\pi_d$ \\
$T_{\max}$ & Maximum refinement depth \\
$a_t$ & Action tuple $(\gamma_t,\beta_t,v_t)$ sampled from $\pi_a$ \\
$\gamma_t,\ \beta_t$ & Scale and shift parameters controlling step size $\alpha_t$ \\
$v_t$ & Normalized refinement direction, $v_t\in\mathbb{R}^{d_v}$ \\
$\alpha_t$ & Effective signed step size, $\alpha_t=f_\alpha(\gamma_t,\beta_t)$ \\
$\pi_d,\ \pi_a$ & Depth controller and action controller \\
$f_{\mathrm{extract}}$ & Projection from $h_0$ to initial reasoning state $s_0$ \\
$g(\cdot)$ & Reasoning-state update function \\
$f_{\mathrm{decode}}$ & Function combining $s_T$ and $h_0$ into the final decoding state \\
$\Delta_t,\ r_t$ & Step-wise likelihood gain and reward \\
$c_d$ & Fixed per-step computation cost \\
$R_t$ & Shaped step return \\
$\beta_{\mathrm{KL}}$ & KL-regularization weight in the shaped return \\
$\lambda$ & Entropy-regularization weight \\
$H$ & Policy entropy \\
$A$ & Advantage estimate, $A=R-\mathbb{E}[R]$ \\
$\mathcal{L}_{\mathrm{RL}},\ \mathcal{L}_{\mathrm{LM}}$ & RL loss and language-modeling loss \\
$\alpha_{\mathrm{RL}}$ & Weight balancing $\mathcal{L}_{\mathrm{LM}}$ and $\mathcal{L}_{\mathrm{RL}}$ \\
\bottomrule
\end{tabular}
\caption{Summary of notation used in the main paper.}
\label{tab:notation}
\end{table}
\twocolumn
\bibliography{example_paper}
\bibliographystyle{icml2025}



\end{document}